\documentclass[cameraready]{Interspeech}

\title{Indic DiarBench: A Multilingual Joint Diarization and ASR Benchmark for Indian Languages}

\author[affiliation={1}, equalcontribution]{Deovrat}{Mehendale}
\author[affiliation={1}, equalcontribution]{Aditya}{Mehndiratta}
\author[affiliation={1},equalcontribution]{Dhruv}{Rathi}
\renewcommand{\authorsep}{\\}
\author[affiliation={2}]{Kaushal}{Bhogale}
\author[affiliation={2}]{Mitesh M.}{Khapra}

\address{
  $^1$ Sarvam AI, India 
  $^2$ AI4Bharat, IIT Madras, India
}

\email{deovrat.mehendale@gmail.com, adityam0309@gmail.com, dhruvsubhashrathi@gmail.com}


\usepackage{comment}
\usepackage{enumitem}
\usepackage{multirow}
\usepackage{indentfirst}
\usepackage{colortbl}
\usepackage{booktabs}
\usepackage{amssymb}

\definecolor{hA}{rgb}{0.102,0.596,0.314}
\definecolor{hB}{rgb}{0.569,0.812,0.376}
\definecolor{hC}{rgb}{0.996,0.878,0.545}
\definecolor{hD}{rgb}{0.988,0.553,0.349}
\definecolor{hE}{rgb}{0.843,0.188,0.153}
\definecolor{hG}{rgb}{0.851,0.851,0.851}

\newcommand{\cA}[1]{\cellcolor{hA}\textcolor{white}{#1}}
\newcommand{\cB}[1]{\cellcolor{hB}{#1}}
\newcommand{\cC}[1]{\cellcolor{hC}{#1}}
\newcommand{\cD}[1]{\cellcolor{hD}{#1}}
\newcommand{\cE}[1]{\cellcolor{hE}\textcolor{white}{#1}}
\newcommand{\gm}{\cellcolor{hG}{--}}
\newcommand{\rot}[1]{\rotatebox{90}{\textbf{#1}\hspace{0.3em}}}

\begin{document}

\maketitle

\begin{abstract}
  In this work, we introduce Indic DiarBench, a speaker diarization and ASR benchmark dataset spanning all 22 scheduled languages of India. This corpus comprises approximately 108 hours of natural multi-speaker audio from near-field meetings, far-field recordings, and in-the-wild audios. All annotations are human-corrected with time-aligned speaker attributed transcriptions. The dataset captures conversational nuance prevalent in Indian speech, such as English code-mixing, dialectal variation, and frequent speaker overlap. 
  To establish a baseline for joint ASR and diarization capabilities we evaluate leading systems including commercial speech APIs and multimodal large language models.
  Indic DiarBench is released as an open-access resource\footnote{\url{https://huggingface.co/datasets/sarvamai/indic-diarbench}} to advance inclusive, multilingual speech technology research for Indian languages.
\end{abstract}

\section{Introduction}

Recent years have seen significant progress in automatic speech recognition (ASR) for Indian languages. Large scale data collection efforts such as IndicVoices~\cite{javed2024indicvoices} have enabled multilingual ASR systems that begin to cover India's linguistic diversity. However, most of this progress has focused on single speaker speech, while many real world scenarios such as meetings, interviews, panel discussions, and casual conversations involve multiple interacting speakers. In practical multi speaker transcription pipelines, speaker diarization is first used to identify and segment speakers, after which ASR is applied to each segment. 

Despite substantial advances in speaker diarization research~\cite{park2021review}, existing datasets and benchmarks remain heavily concentrated in English and a few other high resource languages, leaving no standardized benchmark for multi speaker speech processing in Indian languages. This gap is particularly consequential because diarization and ASR must operate jointly in real deployments. Evaluating them independently masks important failure modes: ASR performance often degrades sharply when applied to the short, fragmented, and overlapping segments produced by diarization systems. As a result, progress in real world conversational transcription requires benchmarks that evaluate speaker attributed ASR under realistic multi speaker conditions, a capability that remains largely unexplored for Indian languages spanning 22 constitutionally recognized languages and four major language families.

\begin{figure}[t]
  \centering
  \includegraphics[width=0.7\columnwidth]{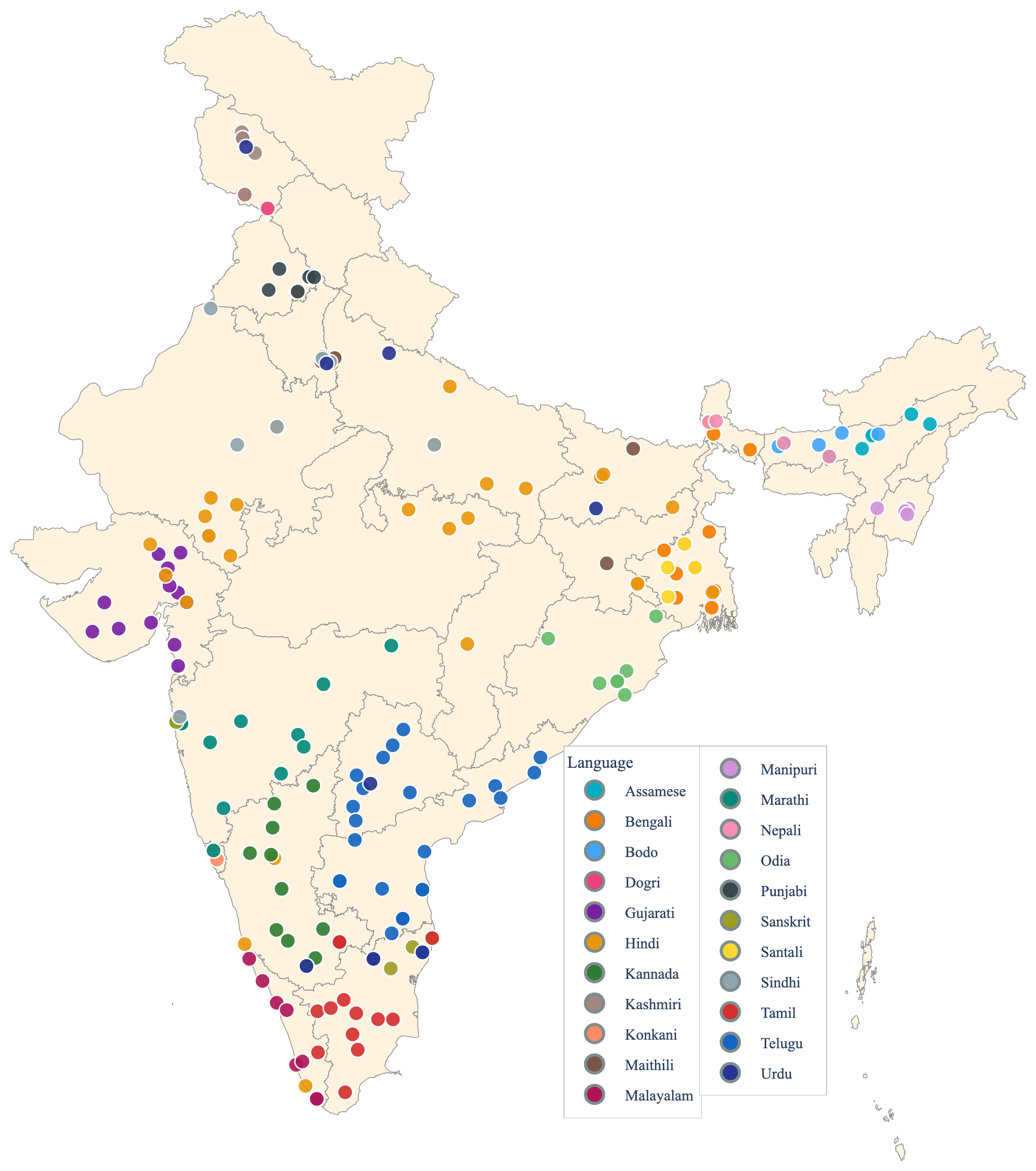}
  \caption{Speaker Data Collection by District and Language}
  \label{fig:ind_map}
\end{figure}

To address this gap, we introduce Indic DiarBench, an open-access benchmark for evaluating speaker-attributed ASR in multilingual Indian conversational speech. Our contributions can be summarised as follows:

\begin{itemize}[nosep,leftmargin=*]
\item A corpus of \textbf{108} hours of conversational speech covering \textbf{all 22 scheduled Indian languages}, collected across near-field meetings, far-field recordings, and in-the-wild YouTube conversations.
\item Meeting recordings include 485 unique speakers from 189 districts across India (Figure \ref{fig:ind_map}). 
The in-the-wild data further includes $\approx 750$ speakers across the 10 most widely spoken Indian languages.
\item \textbf{Human-corrected, speaker-attributed, segment-level transcriptions} paired with aligned speaker time annotations (RTTM format), enabling evaluation of joint diarization and ASR performance.
\item Baseline results from multiple state-of-the-art commercial APIs, multimodal LLMs, and Indic-specialized systems, establishing reference performance across languages and acoustic conditions.

Indic DiarBench, along with all annotations, evaluation protocols, and baseline systems, is publicly released to support reproducible research and advancement of speaker-attributed ASR and diarization for Indian languages.

\end{itemize}

\section{Related Work}
Early diarization benchmarks such as the AMI~\cite{carletta2005ami} and ICSI~\cite{janin2003icsi} meeting corpora provided multi-channel English recordings in controlled rooms. CALLHOME~\cite{callhome} extended coverage to six languages but is limited to telephonic speech with two to six speakers. The DIHARD challenge series~\cite{ryant2021dihard3} pushed evaluation toward degraded and diverse domains, while VoxConverse~\cite{chung2020voxconverse} introduced in-the-wild YouTube audio with audio-visual annotation. For Mandarin, AliMeeting~\cite{yu2022m2met} and AISHELL-4~\cite{fu2021aishell4} provide meeting-style corpora. LibriCSS~\cite{chen2020libricss} targets continuous speech separation with controlled overlap ratios. More recently, SDBench~\cite{durmus2025sdbench} unified 13 datasets under a single evaluation framework, and NOTSOFAR-1~\cite{vinnikov2024notsofar} introduced realistic meeting data with tcpWER evaluation. Despite this progress, none of these benchmarks provide substantial coverage of Indian languages. Table~\ref{tab:diarization_datasets} summarizes key datasets.
At a broader multilingual ASR level, Common Voice~\cite{ardila2020commonvoice} and FLEURS~\cite{conneau2022fleurs} significantly expanded language coverage, but both are primarily designed for single-speaker recognition and do not provide speaker-attributed conversational diarization labels.

For Indian languages, the DISPLACE challenges~\cite{baghel2023displace, kalluri2024displace2} represent the first attempts at diarization benchmarks. DISPLACE~2023 provided 32 hours of conversational audio across seven languages, while the 2024 edition expanded to 158 hours (38 hours labelled). However, the ASR track in 2024 was evaluated on a separate 12-hour subset of cleaner near-field, single-speaker audio. By decoupling diarization from ASR and covering only a subset of Indian Languages, DISPLACE leaves critical gaps for evaluation of speaker-attributed transcriptions for Indian languages in multi-speaker conversational settings.

\begin{table}[t]
  \centering
  \setlength{\tabcolsep}{2pt}
  \caption{Comparison of diarization evaluation datasets. Indic DiarBench is the first to cover all 22 Indian scheduled languages with joint ASR\,+\,diarization labels.}
  \label{tab:diarization_datasets}
  \begin{tabular}{l l c l c c}
    \toprule
    \textbf{Dataset} & \textbf{Lang.} & \textbf{Hrs} & \textbf{Domain} & \textbf{Spkrs} & \textbf{ASR} \\
    \midrule
    CALLHOME         & 6 langs     & 20   & Telephone     & 2--6  & \checkmark \\
    AMI              & English     & 100  & Meeting       & 3--5  & \checkmark \\
    DIHARD III       & Multi       & 33   & Mixed         & 1--10 & --         \\
    VoxConverse      & Multi       & 70 & YouTube       & 1--21 & --         \\
    LibriCSS         & English     & 10   & Simulated     & 8     & \checkmark \\
    AliMeeting       & Mandarin    & 118  & Meeting       & 2--4  & \checkmark \\
    NOTSOFAR-1       & English     & 28   & Distant mtg.  & 4--8  & \checkmark \\
    DISPLACE '24     & 5 Indic     & 38  & Conversation  & 3--5  & -- \\
    \midrule
    \textbf{Ours}    & \textbf{22 Indic} & \textbf{${\sim}$108} & \textbf{Mixed} & \textbf{2--9} & \checkmark \\
    \bottomrule
  \end{tabular}

  \vspace{0.4em}
\end{table}

\section{The Indic DiarBench Corpus}

Indic DiarBench is a multilingual conversational speech benchmark designed to evaluate speaker attributed ASR in realistic multi speaker settings for Indian languages. 
The following subsections describe the data collection process, annotation pipeline, and dataset statistics.

\subsection{Data Collection}
We now describe the recording conditions represented in the corpus, followed by the criteria used for audio selection and the characteristics of the speaker population.

\subsubsection{Recording Conditions}

\begin{itemize}[nosep,leftmargin=*]

\item \textbf{Near-field meetings} (${\sim}$\,\textbf{53}\,hrs): {Recorded using one close-proximity microphone per speaker and designed to capture spontaneous interaction, frequent interjections, and overlapping speech. Participants were not co-located; they joined virtually via an online meeting platform. This setup allowed accurate capture of speaker turns by combining individual microphone streams.} This subset covers all 22 scheduled Indian languages. The top eight languages by native-speaker population (Hindi, Bengali, Marathi, Telugu, Tamil, Gujarati, Punjabi, Kannada) contribute ${\sim}$\,4 hours each, while the remaining 14 languages contribute ${\sim}$\,1.5 hours each.

\item \textbf{Far-field meetings} (${\sim}$\,\textbf{27}\,hrs): Recorded using distant microphones, introducing reverberation, background noise, and variable speaker-to-microphone distances. This subset includes approximately 1.4 to 4.2 hours per language for the top eight languages, with sessions involving 2--8 speakers.

\item \textbf{In-the-wild audio} (${\sim}$\,\textbf{28}\,hrs): Curated from publicly available YouTube videos to capture unconstrained acoustic environments. Approximately 2 hours per language are collected for the ten most widely spoken Indian languages. Selected clips emphasize sustained multi-speaker interaction and avoid broadcast-style content dominated by a single speaker.

\end{itemize}

\subsubsection{Data Curation and Quality Control}

To encourage natural conversation and avoid scripted turn-taking, participants were provided debate topics and quiz questionnaires in advance and asked to speak freely. An initial warm-up portion of each recording is discarded to retain only spontaneous interaction. Sessions with same-gender speakers are included to increase speaker confusability. After collection, all recordings undergo a curation and quality control process to ensure suitability for multi-speaker evaluation.

For meeting recordings, trained language experts review sessions for recording quality, vocabulary diversity, speaker overlap, and conversational spontaneity. Sessions that are overly scripted or exhibit poor audio quality are discarded. Noise cancellation is kept at ``low'' whenever possible during recording to preserve natural ambient sound and conversational artifacts.

For in-the-wild audio, clips are curated from publicly available videos that (\textit{i}) contain sustained multi-speaker interaction, (\textit{ii}) avoid advertisements and non-speech-dominant music, and (\textit{iii}) provide a clear visual indication of the active speaker, enabling reliable speaker identification during annotation.

\subsubsection{Speaker Population}

Following the data collection principles established in IndicVoices~\cite{javed2024indicvoices}, we recruit speakers from diverse geographic, demographic, and dialectal backgrounds across India. Figure~\ref{fig:ind_map} illustrates the geographic distribution of speaker data collection across districts and languages. 
Meeting recordings across near- and far-field conditions include \textbf{485} unique speakers drawn from urban and rural populations across \textbf{22 languages} and \textbf{189 districts}, capturing substantial accent and dialectal variation. Speaker identities and associated metadata (e.g., language, region, and demographics) are tracked using unique speaker IDs and session-level metadata files that will be released with the meetings dataset. The speaker pool is gender-balanced and spans a range of educational backgrounds.

For in-the-wild audio, speaker uniqueness is enforced by restricting the dataset to one video per channel. Speaker embeddings are subsequently clustered to identify potential overlaps across videos, followed by manual verification to ensure distinct speaker identities.

\subsection{Annotation Pipeline}

All recordings are annotated using a unified human-in-the-loop pipeline designed to produce high-quality speaker-attributed transcriptions. The pipeline combines automatic transcription with three stages of human verification to ensure accurate transcriptions, timestamps, and speaker labels.
\begin{enumerate}[nosep,leftmargin=*]

\item \textbf{Bootstrap Transcription.}
Initial transcripts are generated using multiple independent ASR systems, including publicly available models and closed source APIs. Annotators are presented with these hypotheses as editable drafts, reducing annotation effort while limiting bias from any single system.

\item \textbf{Human Transcription and Speaker Attribution.}
Professional annotators produce time-aligned, speaker-attributed transcriptions by verifying and correcting word sequences, timestamps, and speaker labels. No machine-generated annotation is retained without human validation. For meeting recordings, where the number of speakers is known and individual microphone channels provide reliable speaker timing, annotators may adjust timestamps or speaker labels but cannot introduce new speakers. For in-the-wild audio, where visual cues are available, annotators may add, merge, or remove speaker labels when necessary.

\item \textbf{Code-Mixed Transcription.}
Indian conversational speech has frequent code-mixing with English. To accommodate this, annotators produce two transcription formats. The first is a native-script form in which all text, including English words, is rendered in the Indic script. The second is a normalized form in which English words are written using Roman script and numerals using Arabic digits. Both formats are accepted during word error rate computation to avoid penalizing models that differ in output conventions.

\item \textbf{Quality Control.}
Dedicated quality checkers (2–3 per language) verify transcription consistency, code-mixing conventions, numeral representations, non-speech tags (\texttt{<laughter>}, \texttt{<noise>}, \texttt{<cry>}, etc.), and speaker timestamps and labels. Particular attention is paid to overlapping speech segments, which often require multiple rounds of review. Minor corrections are applied at this stage, while major issues are returned to annotators for revision.

\item \textbf{Expert Review.}
Finally, each file undergoes language specific quality checks by an in-house expert (\textbf{superchecker}). Rather than correcting instances directly, the expert identifies systematic issues and returns substandard annotations for revision until quality standards are met.

\end{enumerate}

\subsection{Dataset Statistics}

Indic DiarBench contains approximately \textbf{108} hours of annotated audio across all 22 scheduled Indian languages. Table~\ref{tab:dataset_stats} presents per-language statistics including duration by acoustic condition, speaker counts, and overlap ratios.

\begin{table}[t]
  \centering
  \setlength{\tabcolsep}{3pt}
  \caption{Per-language statistics for Indic DiarBench. All durations are in hours. Hours are broken down by acoustic condition (NF = near-field, FF = far-field, ITW = in-the-wild). Overlap \% is averaged over all conditions. ``--'' indicates that the condition is not available for that language.}
  \label{tab:dataset_stats}
  \resizebox{\columnwidth}{!}{
  \begin{tabular}{l l c c c c c}
    \toprule
    \textbf{Language} & \textbf{Family} & \textbf{NF} & \textbf{FF} & \textbf{ITW} & \textbf{Total} & \textbf{Overlap\,\%} \\
    \midrule
    Assamese    & Indo-Aryan     & 1.5 & --            & --            & 1.5 & 13.9 \\
    Bengali     & Indo-Aryan     & 4.4 & 4.1 & 4.1 & 12.6 & 7.8 \\
    Bodo        & Sino-Tibetan   & 1.6 & --            & --            & 1.6 & 15.2 \\
    Dogri       & Indo-Aryan     & 1.4 & --            & --            & 1.4 & 24.2 \\
    Gujarati    & Indo-Aryan     & 4.1 & 4.2 & 2.8 & 11.1 & 7.6 \\
    Hindi       & Indo-Aryan     & 4.2 & 4   & 2.5 & 10.7 & 16.6 \\
    Kannada     & Dravidian      & 3.7 & 1.4 & 3.3 & 8.5 & 15.2 \\
    Kashmiri    & Indo-Aryan     & 1.1 & --            & --            & 1.1 & 21.2 \\
    Konkani     & Indo-Aryan     & 1.6 & --            & --            & 1.6 & 14.0 \\
    Maithili    & Indo-Aryan     & 1.3 & --            & --            & 1.3 & 24.7 \\
    Malayalam   & Dravidian      & 1.3 & --            & 2.4 & 3.7 & 12.9 \\
    Manipuri    & Sino-Tibetan   & 1.5 & --            & --            & 1.5 & 20.6 \\
    Marathi     & Indo-Aryan     & 4.2 & 3.5 & 2.7 & 10.4 & 11.3 \\
    Nepali      & Indo-Aryan     & 1.3 & --            & --            & 1.3 & 22.9 \\
    Odia        & Indo-Aryan     & 1.5 & --            & 1.6 & 3.1 & 11.1 \\
    Punjabi     & Indo-Aryan     & 4.3 & 4   & 2.4 & 10.6 & 6.1 \\
    Sanskrit    & Indo-Aryan     & 1.6 & --            & --            & 1.6 & 21.4 \\
    Santali     & Austroasiatic  & 1.6 & --            & --            & 1.6 & 6.5 \\
    Sindhi      & Indo-Aryan     & 1.5 & --            & --            & 1.5 & 16.0 \\
    Tamil       & Dravidian      & 4.2 & 2.5 & 3.2 & 10 & 12.4 \\
    Telugu      & Dravidian      & 3.8 & 3   & 2.5 & 9.3 & 20.4 \\
    Urdu        & Indo-Aryan     & 1.6 & --            & --            & 1.6 & 12.5 \\
    \midrule
    \textbf{Total} & \textbf{4 families} & \textbf{53.2} & \textbf{26.8} & \textbf{27.6} & \textbf{${\sim}$108} & \textbf{12.8} \\
    \bottomrule
  \end{tabular}
  }
\end{table}

\noindent\textbf{Limitations.}
The in-the-wild subset covers 10 of the 22 languages; extending coverage to all languages is planned. We do not provide speaker IDs for the in-the-wild subset. The dataset is designed for evaluation rather than training.

\begin{table}[t]
  \centering
  \setlength{\tabcolsep}{2pt}
  \caption{Duration-weighted aggregate metrics across all three acoustic conditions (\%). Best values in \textbf{bold}.}
  \label{tab:overall_results}
  \resizebox{\columnwidth}{!}{
  \begin{tabular}{l l c c c c c c}
    \toprule
    \textbf{Category} & \textbf{Model} & \textbf{DER} & \textbf{cpWER} & \textbf{WDER} & \textbf{Miss} & \textbf{FA} & \textbf{Conf} \\
    \midrule
    \multirow{1}{*}{Indic-Spec.}
      & Sarvam         & \textbf{16.0} & \textbf{38.8} & 33.1 &  \textbf{6.3} & 3.9 &  \textbf{5.9} \\
    \midrule
    \multirow{5}{*}{Comm. APIs}
      & AWS Transcribe    & 23.5 & 43.7 & 34.3 & 13.1 & 3.1 &  7.4 \\
      & ElevenLabs Scribe & 35.0 & 58.3 & 40.7 & 13.6 & 6.2 & 15.3 \\
      & Azure STT         & 34.8 & 60.8 & 39.5 & 24.4 & \textbf{1.7} &  8.7 \\
      & Deepgram Nova-3   & 32.0 & 63.2 & 39.3 & 18.3 & 5.4 &  8.3 \\
      & AssemblyAI        & 40.5 & 88.6 & 43.7 & 25.5 & 5.5 &  9.6 \\
    \midrule
    \multirow{2}{*}{Multimod. LLMs}
      & GPT-4o            & 36.2 & 83.1 & 40.4 & 17.7 & 3.8 & 14.7 \\
      & Gemini~3~Pro      & 74.0 & 58.9 & \textbf{33.0} & 41.7 & 5.8 & 26.5 \\
    \bottomrule
\end{tabular}
  }
\end{table}

\section{Evaluation Setup}

\noindent\textbf{Metrics.} We report evaluation metrics along two complementary axes: acoustic diarization and word-level speaker attribution. For acoustic segmentation, we report Diarization Error Rate (DER) computed without a forgiveness collar and including overlapping speech. 

To jointly evaluate ASR and diarization performance, we use Concatenated minimum-permutation Word Error Rate (cpWER)~\cite{watanabe2020chime6} and Word Diarization Error Rate (WDER)~\cite{shafey2019wder}. cpWER primarily reflects transcription accuracy after resolving speaker permutations, whereas WDER explicitly penalizes speaker attribution errors and therefore more directly captures diarization quality.

\noindent\textbf{Models:} We benchmark several prominent ASR–diarization systems, including commercial speech APIs (Sarvam AI~\cite{sarvam_asr_blog}, Deepgram Nova-3~\cite{deepgram_nova3}, ElevenLabs Scribe~\cite{elevenlabs_stt}, AssemblyAI Universal-2~\cite{assemblyai_universal2}, Azure STT~\cite{azure_stt}, AWS Transcribe~\cite{aws_transcribe}) and multimodal large language models (Gemini 3 Pro~\cite{gemini3_blog} and GPT-4o Transcribe~\cite{gpt4o_transcribe_api}).

Evaluation is restricted to systems capable of producing joint ASR and diarization outputs, as assessing diarization independently of transcription is increasingly misaligned with the requirements of modern speech applications. Consequently, diarization-only models (e.g., Pyannote \cite{Bredin23}) are excluded. Language coverage varies across systems, as not all commercial providers support all 22 scheduled Indian languages.

\begin{figure*}[!t]
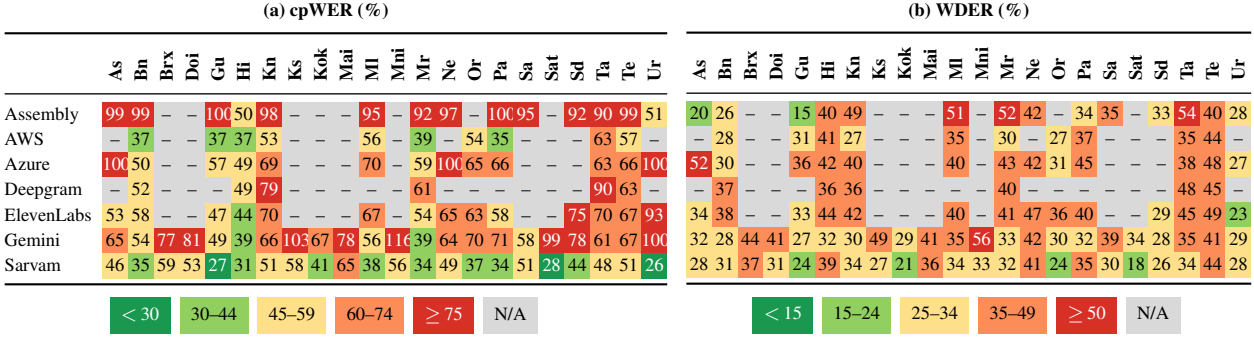

  \centering
  \setlength{\tabcolsep}{0pt} 
  \renewcommand{\arraystretch}{1.25} 

  \begin{minipage}[t]{0.5\textwidth}
    \centering
    {\scriptsize\bfseries (a) cpWER (\%)}\\[0.4em]
    {\scriptsize
    \begin{tabular}{@{}l@{\hspace{4pt}}*{22}{p{1.38em}<{\centering}}@{}}
      \toprule
      & \rot{As} & \rot{Bn} & \rot{Brx} & \rot{Doi} & \rot{Gu} & \rot{Hi} & \rot{Kn} & \rot{Ks} & \rot{Kok} & \rot{Mai} & \rot{Ml} & \rot{Mni} & \rot{Mr} & \rot{Ne} & \rot{Or} & \rot{Pa} & \rot{Sa} & \rot{Sat} & \rot{Sd} & \rot{Ta} & \rot{Te} & \rot{Ur} \\
      \midrule
      Assembly       & \cE{99} & \cE{99} & \gm & \gm & \cE{100} & \cC{50} & \cE{98} & \gm & \gm & \gm & \cE{95} & \gm & \cE{92} & \cE{97} & \gm & \cE{100} & \cE{95} & \gm & \cE{92} & \cE{90} & \cE{99} & \cC{51} \\
      AWS          & \gm & \cB{37} & \gm & \gm & \cB{37} & \cB{37} & \cC{53} & \gm & \gm & \gm & \cC{56} & \gm & \cB{39} & \gm & \cC{54} & \cB{35} & \gm & \gm & \gm & \cD{63} & \cC{57} & \gm \\
      Azure          & \cE{100} & \cC{50} & \gm & \gm & \cC{57} & \cC{49} & \cD{69} & \gm & \gm & \gm & \cD{70} & \gm & \cC{59} & \cE{100} & \cD{65} & \cD{66} & \gm & \gm & \gm & \cD{63} & \cD{66} & \cE{100} \\
      Deepgram       & \gm & \cC{52} & \gm & \gm & \gm & \cC{49} & \cE{79} & \gm & \gm & \gm & \gm & \gm & \cD{61} & \gm & \gm & \gm & \gm & \gm & \gm & \cE{90} & \cD{63} & \gm \\
      ElevenLabs     & \cC{53} & \cC{58} & \gm & \gm & \cC{47} & \cB{44} & \cD{70} & \gm & \gm & \gm & \cD{67} & \gm & \cC{54} & \cD{65} & \cD{63} & \cC{58} & \gm & \gm & \cE{75} & \cD{70} & \cD{67} & \cE{93} \\
      Gemini         & \cD{65} & \cC{54} & \cE{77} & \cE{81} & \cC{49} & \cB{39} & \cD{66} & \cE{103} & \cD{67} & \cE{78} & \cC{56} & \cE{116} & \cB{39} & \cD{64} & \cD{70} & \cD{71} & \cC{58} & \cE{99} & \cE{78} & \cD{61} & \cD{67} & \cE{100} \\
      Sarvam      & \cC{46} & \cB{35} & \cC{59} & \cC{53} & \cA{27} & \cB{31} & \cC{51} & \cC{58} & \cB{41} & \cD{65} & \cB{38} & \cC{56} & \cB{34} & \cC{49} & \cB{37} & \cB{34} & \cC{51} & \cA{28} & \cB{44} & \cC{48} & \cC{51} & \cA{26} \\
      \bottomrule
    \end{tabular}
    }\\[0.4em]
    {\scriptsize
      \colorbox{hA}{\textcolor{white}{\strut\,\textless\,30\,}} \ 
      \colorbox{hB}{\strut\,30--44\,} \ 
      \colorbox{hC}{\strut\,45--59\,} \ 
      \colorbox{hD}{\strut\,60--74\,} \ 
      \colorbox{hE}{\textcolor{white}{\strut\,$\geq$\,75\,}} \ 
      \colorbox{hG}{\strut\,N/A\,}
    }
  \end{minipage}\ignorespaces
  \begin{minipage}[t]{0.5\textwidth}
    \centering
    {\scriptsize\bfseries (b) WDER (\%)}\\[0.4em]
    {\scriptsize
    \begin{tabular}{@{}*{22}{p{1.38em}<{\centering}}@{}}
      \toprule
      \rot{As} & \rot{Bn} & \rot{Brx} & \rot{Doi} & \rot{Gu} & \rot{Hi} & \rot{Kn} & \rot{Ks} & \rot{Kok} & \rot{Mai} & \rot{Ml} & \rot{Mni} & \rot{Mr} & \rot{Ne} & \rot{Or} & \rot{Pa} & \rot{Sa} & \rot{Sat} & \rot{Sd} & \rot{Ta} & \rot{Te} & \rot{Ur} \\
      \midrule
      \cB{20} & \cC{26} & \gm & \gm & \cB{15} & \cD{40} & \cD{49} & \gm & \gm & \gm & \cE{51} & \gm & \cE{52} & \cD{42} & \gm & \cC{34} & \cD{35} & \gm & \cC{33} & \cE{54} & \cD{40} & \cC{28} \\
      \gm & \cC{28} & \gm & \gm & \cC{31} & \cD{41} & \cC{27} & \gm & \gm & \gm & \cD{35} & \gm & \cC{30} & \gm & \cC{27} & \cD{37} & \gm & \gm & \gm & \cD{35} & \cD{44} & \gm \\
      \cE{52} & \cC{30} & \gm & \gm & \cD{36} & \cD{42} & \cD{40} & \gm & \gm & \gm & \cD{40} & \gm & \cD{43} & \cD{42} & \cC{31} & \cD{45} & \gm & \gm & \gm & \cD{38} & \cD{48} & \cC{27} \\
      \gm & \cD{37} & \gm & \gm & \gm & \cD{36} & \cD{36} & \gm & \gm & \gm & \gm & \gm & \cD{40} & \gm & \gm & \gm & \gm & \gm & \gm & \cD{48} & \cD{45} & \gm \\
      \cC{34} & \cD{38} & \gm & \gm & \cC{33} & \cD{44} & \cD{42} & \gm & \gm & \gm & \cD{40} & \gm & \cD{41} & \cD{47} & \cD{36} & \cD{40} & \gm & \gm & \cC{29} & \cD{45} & \cD{49} & \cB{23} \\
      \cC{32} & \cC{28} & \cD{44} & \cD{41} & \cC{27} & \cC{32} & \cC{30} & \cD{49} & \cC{29} & \cD{41} & \cD{35} & \cE{56} & \cC{33} & \cD{42} & \cC{30} & \cC{32} & \cD{39} & \cC{34} & \cC{28} & \cD{35} & \cD{41} & \cC{29} \\
      \cC{28} & \cC{31} & \cD{37} & \cC{31} & \cB{24} & \cD{39} & \cC{34} & \cC{27} & \cB{21} & \cD{36} & \cC{34} & \cC{33} & \cC{32} & \cD{41} & \cB{24} & \cD{35} & \cC{30} & \cB{18} & \cC{26} & \cC{34} & \cD{44} & \cC{28} \\
      \bottomrule
    \end{tabular}
    }\\[0.4em]
    {\scriptsize
      \colorbox{hA}{\textcolor{white}{\strut\,\textless\,15\,}} \ 
      \colorbox{hB}{\strut\,15--24\,} \ 
      \colorbox{hC}{\strut\,25--34\,} \ 
      \colorbox{hD}{\strut\,35--49\,} \ 
      \colorbox{hE}{\textcolor{white}{\strut\,$\geq$\,50\,}} \ 
      \colorbox{hG}{\strut\,N/A\,}
    }
  \end{minipage}

  \caption{Per-language (a) cpWER and (b) WDER (\%) heatmap.}
  \label{fig:cpwer_heatmap}
\end{figure*}
\section{Results and Analysis}
\label{sec:results}

{For evaluation, all systems were provided the same single-channel mixed audio to ensure fairness.} Figure~\ref{fig:cpwer_heatmap} presents per-language cpWER and WDER for all evaluated systems across 22 Indic languages. Grey cells indicate unsupported languages; consequently, global averages over languages are inherently skewed, and we focus instead on duration-weighted aggregates and category-wise trends. Table~\ref{tab:overall_results} summarizes duration-weighted performance across all three acoustic conditions.

\noindent\textbf{Model Comparison.}
We first compare overall performance across model categories. Among all evaluated systems, the Indic-specialized Sarvam pipeline consistently achieves the strongest results, obtaining the lowest DER (16.0\%) and cpWER (38.8\%) in Table~\ref{tab:overall_results}. Other model APIs exhibit moderate performance, with AWS Transcribe performing best in this group (23.5\% DER, 43.7\% cpWER), while other APIs show substantially higher error rates. Multimodal LLMs present a contrasting trade-off: Gemini~3~Pro achieves competitive WDER (33.0\%) due to strong ASR quality on detected speaker segments, but suffers from very high DER (74.0\%), whereas GPT-4o achieves better diarization accuracy but exhibits poor cpWER, particularly on lower-resource languages.

\begin{figure}[t]
  \centering
  \includegraphics[width=1.0\linewidth]{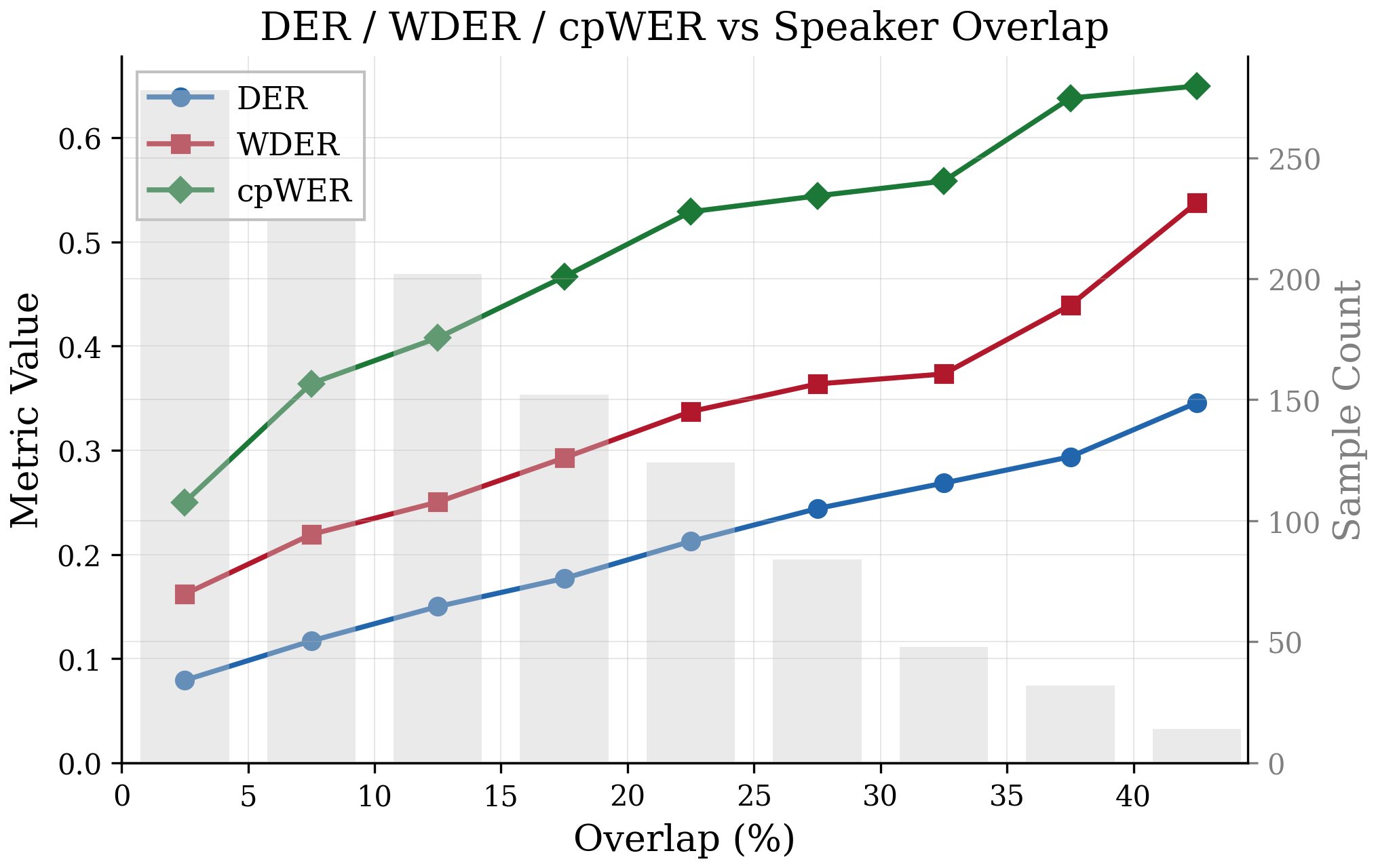}
  \caption{Metrics variation vs overlap ratio}
  \label{fig:overlap_speakers}
\end{figure}

\noindent\textbf{Error Analysis.}
To better understand system behavior beyond aggregate metrics, we decompose DER into missed detection, false alarm, and speaker confusion errors (see Table \ref{tab:overall_results}). This breakdown reveals distinct failure patterns across model classes. Indic-specialized pipelines exhibit a balanced error profile: Sarvam’s 16.0\% DER is composed of 6.3\% missed detection, 3.9\% false alarms, and 5.9\% speaker confusion, indicating no single dominant failure mode. 
In contrast, multimodal LLMs are dominated by missed detection errors. Gemini~3~Pro’s 74.0\% DER includes 41.7\% missed detection and 26.5\% confusion, largely due to unreliable timestamping and missed detection on small utterances like affirmations and interjections. Despite this, its cpWER of 58.9\% remains competitive with commercial ASR APIs, indicating strong transcription quality when speaker segments are correctly identified. Generic Commercial APIs exhibit more model-specific behavior: Azure STT achieves low false alarm rates (1.7\%) due to conservative voice-activity detection, but incurs high missed detection (24.4\%), while AWS Transcribe has relatively more balanced error decomposition.

\noindent\textbf{Effects of overlap}
In the Indic-specific model (Sarvam), across the 22 languages overlap ratio is strongly correlated with DER and cpWER (Figure~\ref{fig:overlap_speakers}).

\noindent\textbf{Performance across languages.}
We provide language-wise and near-field vs far-field results in the supplementary material (DatasetSummary.csv) and discuss them briefly here. In the near-field condition, Telugu emerges as the most challenging language, exhibiting the highest overlap~(24.7\%) and 4.5 speakers per recording on average, yielding a DER of 27.7\%.
Maithili~(24.7\% overlap, 4.7 speakers, 22.7\% DER) and Dogri~(24.2\%, 5.3 speakers, 23.2\% DER) follow closely.
At the other end, Santali~(6.5\% overlap, 4.2 speakers, 9.7\% DER) and Urdu~(12.5\%, 3.8 speakers, 12.8\% DER) are among the easiest; these trends are consistently reflected across all evaluated systems.
In-the-wild YouTube recordings, which have the lowest overlap~(6.5\%) and cleaner turn-taking, yield the best performance.
Lower-resource languages compound these effects.
The 12 languages available only in near-field recordings show higher DER and cpWER compared to the 10 higher-resource languages that span multiple acoustic conditions.
Across language families, Dravidian languages~(Kannada, Malayalam, Tamil, Telugu) show near-field cpWER roughly 5 percentage points above Indo-Aryan languages at comparable DER.

\section{Conclusion}

We present \textbf{Indic DiarBench}, the first open benchmark for joint diarization and speaker-attributed ASR spanning all 22 scheduled Indian languages. By unifying near-field meetings, far-field recordings, and in-the-wild conversations, the benchmark captures realistic variation in speaker counts, overlap ratios, and acoustic conditions, establishing a standardized evaluation framework for multi-speaker conversational speech in Indian languages. Our results underline that robust speaker-attributed recognition remains challenging in short-utterance, high-overlap, and low-resource settings, motivating continued research on tightly coupled diarization and ASR systems.

\ifcameraready
  \section{Acknowledgments}
  {We thank Sshubam, Sadakopa, and Vamsi from Sarvam AI for generously giving their time and helping with the YouTube data collection effort. We also thank the language experts at Sarvam AI and AI4Bharat for their excellent work; this effort would not have been possible without their contributions.}
\fi

\section{Generative AI use disclosure}
Generative AI tools were used only for limited language editing and polishing of parts of the manuscript. All technical content, analyses, results, and conclusions were produced and verified by the authors.

\bibliographystyle{IEEEtran}
\bibliography{mybib}
\end{document}